\documentclass[journal]{IEEEtran}
\usepackage{times}  
\usepackage{helvet}  
\usepackage{courier}  
\usepackage{url}  
\usepackage{graphicx}  
\usepackage{subfigure}
\usepackage{colortbl,booktabs}
\usepackage{float}
\usepackage{multirow}

\usepackage{amsmath,amsfonts}
\usepackage{amsthm}
\usepackage{amssymb,amsopn}

\usepackage{graphicx}
\usepackage{textcomp}
\usepackage{bm}
\usepackage{multirow}
\usepackage[ruled]{algorithm2e}
\usepackage{wrapfig}
\usepackage[graphicx]{realboxes}
\usepackage{graphics}
\usepackage{epsfig}
\usepackage{url}
\usepackage{graphicx}
\usepackage{makecell,rotating}
\usepackage{xspace}
\usepackage{hyperref}
\usepackage{marginnote}
\newcommand{\vct}[1]{\boldsymbol{#1}} 
\newcommand{\mat}[1]{\boldsymbol{#1}} 
\newcommand{\ie}{\emph{i.e.}\xspace} 
\newcommand{\etal}{\emph{et al.}\xspace} 

\newcommand{\eg}{\emph{e.g.}\xspace}

\newcommand{\T}{^{\textrm T}} 

\usepackage{lineno}
\usepackage{framed,multirow}
\usepackage{cite}
\usepackage{graphicx}
\usepackage{longtable}
\usepackage{bm}
\usepackage{epsf, epsfig}
\usepackage{epstopdf}
\usepackage{subfigure}
\usepackage{color}
\usepackage{multirow}
\usepackage{ulem}
\usepackage{diagbox}
\usepackage{footnote}
\usepackage{booktabs}
\usepackage{bbding}
\makesavenoteenv{tabular}
\makesavenoteenv{table}

\usepackage{xargs}
\usepackage[pdftex,dvipsnames,table]{xcolor} 

\usepackage{marginnote}

\DeclareMathOperator*{\argmin}{arg\,min}
\ifCLASSINFOpdf
\else
\fi



\newcommand{\infor}[2][]{{%
 \let\marginpar\marginnote
 \reversemarginpar\info[#1]{{\bf#2}}}}

\usepackage{amsmath}

\begin{document}

\title{Meta Ordinal Regression Forest for Medical Image Classification with Ordinal Labels}

\author{
        Yiming~Lei,
        Haiping~Zhu,
        Junping~Zhang,~\IEEEmembership{Member,~IEEE,}
        Hongming~Shan,~\IEEEmembership{Member,~IEEE}

\thanks{Y. Lei, H. Zhu, and J. Zhang are with the Shanghai Key Laboratory of Intelligent Information Processing, School of Computer Science, Fudan University, Shanghai 200433, China (e-mail: ymlei@fudan.edu.cn; hpzhu14@fudan.edu.cn; jpzhang@fudan.edu.cn)}
\thanks{H. Shan is with the Institute of Science and Technology for Brain-inspired Intelligence and MOE Frontiers Center for Brain Science, Fudan University, Shanghai 200433, China, and also with the Shanghai Center for Brain Science and Brain-inspired Technology, Shanghai 200031, China (e-mail: hmshan@fudan.edu.cn)}
\thanks{\emph{Corresponding author}: Hongming Shan.}
}

\maketitle


\begin{abstract}
The performance of medical image classification has been enhanced by deep convolutional neural networks (CNNs), which are typically trained with cross-entropy (CE) loss. However, when the label presents an intrinsic ordinal property in nature, \eg, the development from benign to malignant tumor, CE loss cannot take into account such ordinal information to allow for better generalization. To improve model generalization with ordinal information, we propose a novel meta ordinal regression forest (MORF) method for medical image classification with ordinal labels, which learns the ordinal relationship through the combination of convolutional neural network and differential forest in a meta-learning framework. The merits of the proposed MORF come from the following two components:
a tree-wise weighting net (TWW-Net) and a grouped feature selection (GFS) module. First, the TWW-Net assigns each tree in the forest with a specific weight that is mapped from the classification loss of the corresponding tree. Hence, all the trees possess varying weights, which is helpful for alleviating the tree-wise prediction variance. 
Second, the GFS module enables a dynamic forest rather than a fixed one that was previously used, allowing for random feature perturbation. During training, we alternatively optimize the parameters of the CNN  backbone and TWW-Net in the meta-learning framework through calculating the Hessian matrix. Experimental results on two medical image classification datasets with ordinal labels, \ie, LIDC-IDRI and Breast Ultrasound Dataset, demonstrate the superior performances of our MORF method over existing state-of-the-art methods.
\end{abstract}


\IEEEpeerreviewmaketitle

\section{Introduction}
\label{sec:introduction}
\begin{figure}[t]
\centering
\includegraphics[width=1\linewidth]{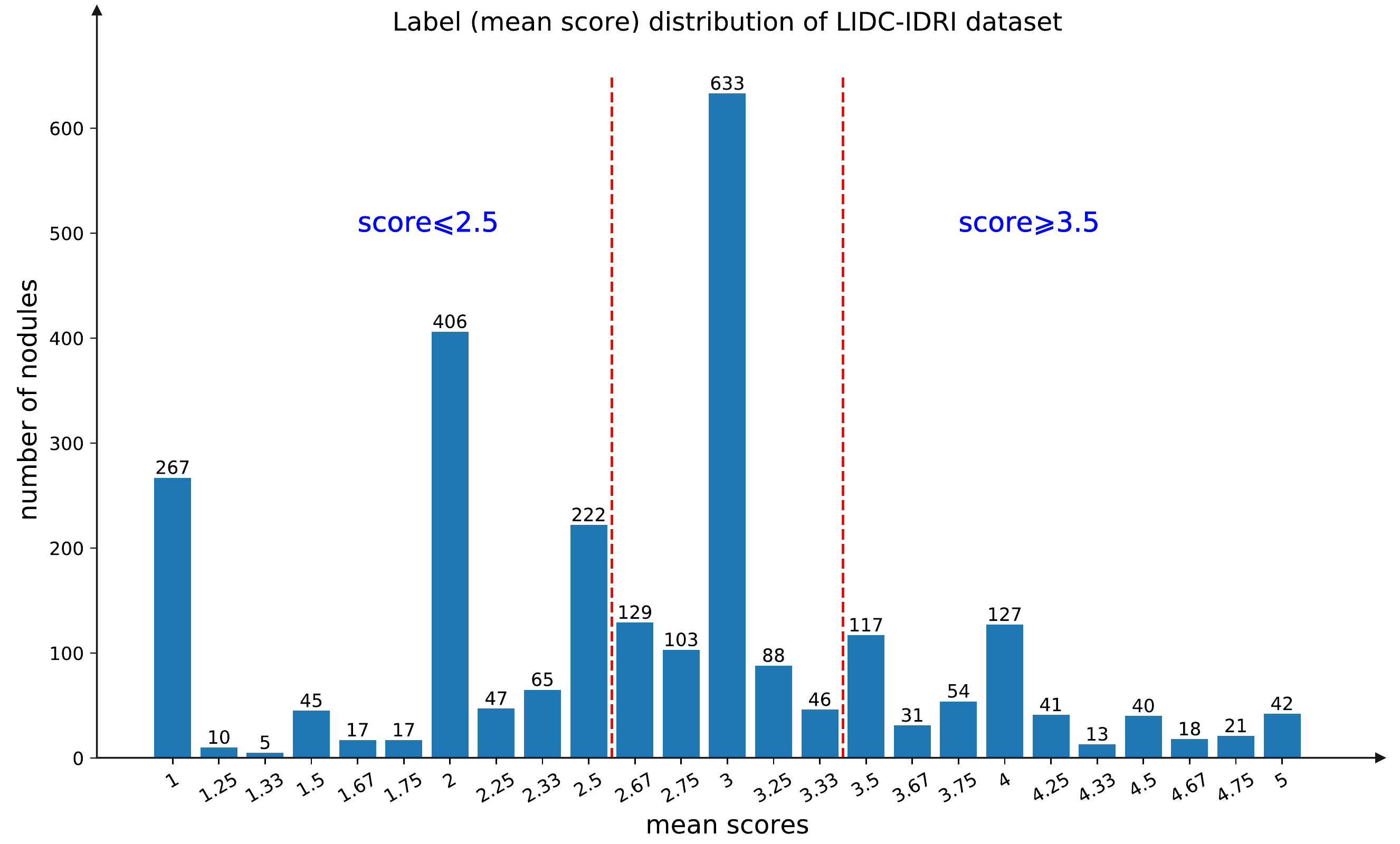}
\caption{The histogram of averaged malignant scores of nodules in LIDC-IDRI dataset. The red dashed lines split the nodules into three groups; \ie, benign ($\leq 2.5$), unsure ($2.5 \sim 3.5$), and malignant ($\ge 3.5$).}
\label{fig:distribution}
\end{figure}

\IEEEPARstart{M}{edical} image classification has been assisted by the deep learning technique~\cite{ResNet,shen2019deep,deeplung,shen2018deep,kontschieder2015deep,zhang2020scis,wang2020scis,lei2020shape} and has achieved tremendous progress in the past decade. Early detection and treatment of some diseases, such as cancers, are critical for reducing mortality. Fortunately, it is implicit in medical images that the image information across the different clinical stages exhibits an ordinal relationship, which can be used to improve model generalization. For example, computed tomographic (CT) images of lung nodule~\cite{LIDC} are given with the malignancy scores from 1 to 5, where 1 means highly unlikely to be malignant, 3 is indeterminate, and 5 is highly likely to be malignant. The majority of existing lung nodule classification methods conduct binary classification while discarding indeterminate or unsure nodules~\cite{deeplung,lei2020shape,fully3d2017,2017tumornet,multiscale_cnn,2017multilevel3d,2016multiview}. In other words, the unsure nodules that are between benign and malignant and cannot be classified by radiologists based on current scans become useless~\cite{wu2019learning}. As shown in Fig.~\ref{fig:distribution}, a large number of nodules are indeterminate and then discarded in the binary classification problem. It is evident that the images with ordinal labels represent the development of the lesions, as do other diseases such as breast cancer. Considering that deep learning methods are data-hungry and that these medical images differ from natural images in fewer discriminative patterns, \emph{leveraging the ordinal relationship among limited medical data for training deep learning models} is becoming an important topic. 

Generally, most medical image classification methods work by feeding medical images into convolutional neural networks (CNNs) and updating the parameters of the CNNs based on cross-entropy (CE) loss. However, CE loss is inferior for fitting the ordinal distribution of labels. Therefore, ordinal regression-based methods have been explored for medical image classification with ordinal labels~\cite{wu2019learning,gutierrez2015ordinal,frank2001simple}. A simple solution is to construct a series of binary classification problems and evaluate the cumulative probabilities of all binary classifiers~\cite{frank2001simple,gutierrez2015ordinal}. However, those binary classifiers are trained separately, which ignores the ordinal relationship. Recently, the unsure data model (UDM)~\cite{wu2019learning},  neural-stick breaking (NSB)~\cite{liu2018ordinal}, unimodal method~\cite{beckham2017unimodal}, and soft ordinal label (SORD)~\cite{diaz2019soft} have been proposed to improve ordinal regression performance based on rectified label space or the probability calculation of the output of the CNNs.

Another method of ordinal regression without changing target distributions is based on the combination of random forest and CNNs, which have been evaluated to successfully estimate human age using facial images~\cite{shen2018deep,kontschieder2015deep,shen2017label}. These models regard the largest probability among all the dimensions of the learned tree-wise distributions as the final prediction. To incorporate the global ordinal relationship with forest-based methods, Zhu~\etal proposed the convolutional ordinal regression forest (CORF) to allow the forest to predict the ordinal distributions~\cite{zhu2021convolutional}. However, these forest-based methods suffer from the following two drawbacks: 1) the compositions of all trees depend on the random selection of split nodes from the feature vector of the fully-connected (FC) layer, and the structure of the constructed forest is fixed at the very beginning of training, leading to poor generalization due to the lack of the random perturbation of features, as suggested in~\cite{breiman2001random}; and 2) there exists the tree-wise prediction variance (tree-variance) because the final prediction of the forest is the average of the results obtained by all trees, \ie, all trees share the same weights and contribute equally to the final prediction.

\begin{figure*}[t]
\centering
\includegraphics[width = 1\linewidth]{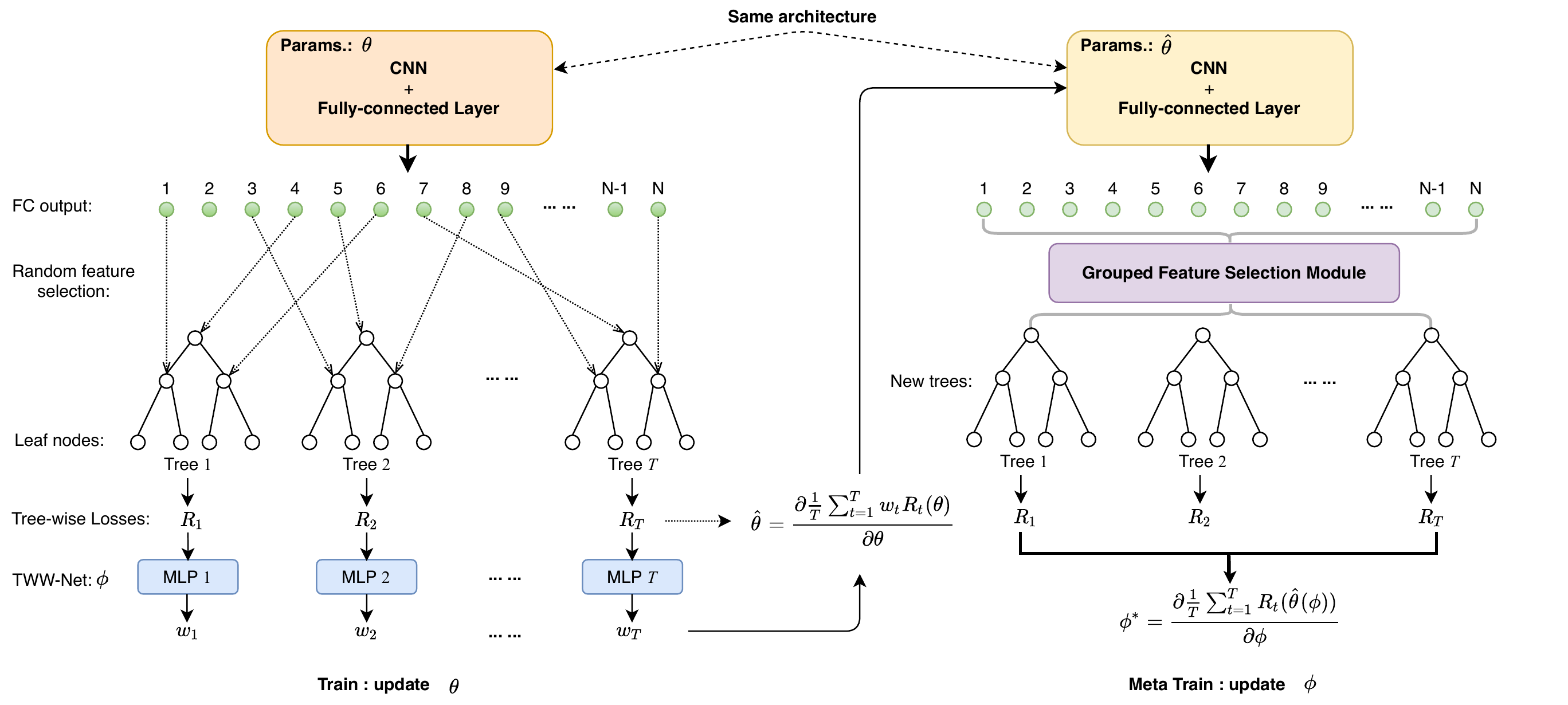}
\caption{The proposed MORF framework. \textbf{Left}: the deep ordinal regression forest with random construction of forest, which is followed by the TWW-Net. \textbf{Right}: the deep ordinal regression forest followed by the TWW-Net, where the forest is constructed from the GFS. During the meta train stage, the GFS features are used to guide the update of $\vct{\theta}$. The MORF framework involves three parts of parameters: $\vct{\theta}$ (CNN), $\vct{\pi}$ (leaf nodes, \ie, the ordinal distributions), and $\vct{\phi}$ (TWW-Net). Note that the parameter $\mat{\widehat{\theta}}$ on the right side is the first-order derivative calculated via~\eqref{eq:theta_hat}. And $\mat{\widehat{\theta}}$ is used to obtain the TWW-Net parameter $\mat{\phi}^{(u+1)}$ through~\eqref{eq:update_phi}. 
}
\label{fig:framework}
\end{figure*}

To address the aforementioned problems, we propose a meta ordinal regression forest (MORF) for medical image classification with ordinal labels. 
Fig.~\ref{fig:framework} shows the overall framework of MORF including three parts: a CNN backbone parameterized by $\mat{\theta}$ to extract feature representation from the input medical image, a tree-wise weighting network (TWW-Net) parameterized by $\mat{\phi}$ to learn the tree-wise weights for reduced tree-wise prediction variance, and a grouped feature selection (GFS) module to construct the dynamic forest that is equipped with random feature perturbation. We adopt the meta-learning framework to optimize the parameters $\mat{\theta}$ and $\mat{\phi}$ alternatively~\cite{shu2019meta,liu2019self}. The way we use meta-learning is similar to those in~\cite{shu2019meta,liu2019self}, which calculates the second derivatives, \ie Hessian matrix, via meta data or meta tasks. The main difference is that our meta data are selected features via the GFS module rather than the original images; the meta data in~\cite{shu2019meta} are a subset of the validation set that contains equal numbers of images of all classes, which may be infeasible in medical cases because some classes have fewer samples. With meta-learning optimization, CNN parameters can be updated with the guide of the dynamic forest, which can achieve better generalizability.

The main contributions of this paper are summarized as follows.
\begin{itemize}
	\item We propose a meta ordinal regression forest (MORF) for medical image classification with ordinal labels, which enables the forest on top of the CNNs to maintain the random perturbation of features. The MORF comprises a CNN backbone, a TWW-Net, and a GFS module.
	\item TWW-Net assigns each tree in the forest with a specific weight and alleviates the tree-wise variance that exists in the previous deep forest-related methods. Furthermore, we provide a theoretical analysis of the weighting scheme of TWW-Net and demonstrate how the meta data can guide the learning of the backbone network. 
	\item The GFS module only works in the meta training stage for generating the dynamic forest that is incorporated the feature random perturbation. Combined with the TWW-Net, the final trained model can be further enhanced through the randomness of the dynamic forest.
	\item The experimental results of two medical image classification datasets with ordinal labels demonstrate the superior performance of MORF over existing methods including classification and ordinal regression. Furthermore, we also verify that our MORF can enhance the benign-malignant classification when leveraging the unsure data into the training set on the LIDC-IDRI dataset.
\end{itemize}

We note that this work extends our previous conference paper~\cite{lei2020meta} with the following major improvements:
\begin{enumerate}
	\item We further explore the relationship between our re-constructed dynamic forest and the random forest, which unveils that our MORF has stronger generalizability than the previous deep forest methods due to its retained random perturbation during training.
	\item We improve GFS module by using random selection without replacement, where the resulting dimension of FC output should be equal to the number of split nodes in a forest. This improvement can avoid selecting unused elements during meta-training stage and achieve further performance gain.
	\item We provide the detailed training algorithm and the theoretical analysis on the meta weighting scheme of TWW-Net.
	\item For the LIDC-IDRI dataset, we demonstrate that the binary classification on the benign and malignant classes can be improved by MORF. Most importantly, when we added the unsure data into the training set, the binary classification results on the test set are improved further.
	\item We conduct extra experiments on a new BUSI dataset~\cite{al2020dataset} to evaluate the performance of the methods.
\end{enumerate}

\section{Related Work}
\label{sec:related_work}
In this section, we review the related work on the following three aspects: 1) CNN-based methods for medical image classification with emphasis on lung nodule classification and breast cancer diagnosis, 2) ordinal regression methods, and 3) meta-learning methods.

\subsection{CNN-based Medical Image Classification}
Medical image classification, such as lung nodule and breast tumor classification, has benefited from advanced CNN architectures and learning strategies. In this paper, we focus on medical image classification with ordinal labels.

\noindent\textbf{Lung nodule classification.}\quad Liu~\etal combined statistical features and artificial neural networks to detect lung nodules in full-size CT images~\cite{liu2017cade}. Shen~\etal applied a single column network to classify lung nodule images with different sizes~\cite{multiscale_cnn}. Dou~\etal~\cite{2017multilevel3d} explored an ensemble of subnetworks, each of which has a specific convolutional kernel size. Cao~\etal trained two 3-D networks on original data and augmented data, and combined them for lung nodule detection~\cite{BIBM3}. However, 3-D networks are difficult to train with limited medical data~\cite{fang20193d,muzahid2020curvenet}. Another kind of method combines low- and high-level features that come from U-Net-like network architectures~\cite{lei2020shape,unet,shan20183d,ibtehaz2020multiresunet,huang2020unet,wu2021pneumothorax}, which can avoid information loss through the concatenation of different features.

\noindent\textbf{Breast cancer diagnosis.}\quad Breast cancer diagnosis has also been enhanced by deep learning~\cite{shen2019deep,liu2018ordinal,ciritsis2019automatic,nascimento2016breast,wang2020breast,sun2020deep,hu2020deep,chiao2019detection,wu2019deep,akselrod2019predicting,dhungel2016automated}. In~\cite{hu2020deep,hagos2018improving}, the authors conducted the CNN-based image fusion, feature fusion, and classifier fusion methods to classify breast tumors in ultrasound (US) images. Wu~\etal explored the binary classification (benign and malignant) of breast cancer based on the proposed view-wise, image-wise, breast-wise, and joint learning approaches~\cite{wu2019deep}. Ayelet~\etal combined CNN learned features with the electronic health records to improve the early detection of breast cancer using mammograms~\cite{akselrod2019predicting}. Dhungel~\etal applied a pretrained CNN model in the classification of breast cancer and verified the effectiveness of deep features against traditionally hand-crafted features~\cite{dhungel2016automated}. Hagos~\etal incorporated symmetry patches with the symmetric locations of positive tumor patches to help improve breast cancer classification performance~\cite{hagos2018improving}.

All the methods we reviewed above did not take into account the progression of the diseases that implies the intrinsic ordinal relationship among the classes. In this paper, we use widely-used backbone networks such as VGG~\cite{vgg} and ResNet~\cite{ResNet} to verify the effectiveness of our method in exploring ordinal relationship.

\subsection{Ordinal Regression}
Ordinal regression is a classical problem that predicts ordinal labels~\cite{gutierrez2015ordinal,frank2001simple,zhu2020scis}, such as facial age~\cite{zhu2021convolutional}, aesthetic image classification~\cite{guo2020large}, and medical image malignancy ranking~\cite{wu2019learning,liu2018ordinal,beckham2017unimodal}. Beckham~\etal enforced each element of the FC output to obey unimodal distributions such as Poisson and Binomial~\cite{beckham2017unimodal}. The unimodal method surpassed the normal cross-entropy baseline. Neural stick-breaking (NSB) was proposed in~\cite{liu2018ordinal}, whose output is a $(N-1)$-dimensional vector representing $N-1$ boundaries where $N$ is the number of classes. NSB guaranteed that the cumulative probabilities will monotonically decrease. The unsure data model (UDM) is a strategy-driven method and focuses more on that the fact that normal samples and disease samples should be classified with high precision and high recall, respectively~\cite{wu2019learning}. The UDM incorporates some additional parameters associated with techniques like ordinal regression, class imbalance, and their proposed strategies. Although the UDM outperforms the unimodal method and NSB, it requires more effort to tune the model to obtain the optimal additional parameters. Soft ordinal label (SORD) converted the ground truth labels of all classes in a soft manner by penalizing the distance between the true rank and the label value~\cite{diaz2019soft}.

The methods discussed above contain a DNN backbone followed by a modified classifier (FC layer) except for the SORD method. This pipeline cannot avoid the redundant use of the FC feature and may lead to overfitting. Recently appeared deep random forest-based methods targeted this problem. The deep neural decision forest (DNDF) defines a differentiable partition function for each split node~\cite{kontschieder2015deep}. Hence, the forest can be updated jointly with the deep networks through backpropagation. The label distribution learning forest (LDLF) extended DNDF to output a predicted distribution~\cite{shen2017label}. However, the DNDF and LDLF have difficulty guaranteeing the global ordinal relationship of the predictions of leaf nodes. Convolutional ordinal regression forest (CORF) was proposed to incorporate the constraint of ordinal relation in the loss function, which enabled the output of the forest to be globally ordinal (monotonically decrease)~\cite{zhu2021convolutional}.

In this paper, our MORF further improves the efficiency of the FC feature and incorporates random perturbation of features for the forest. Moreover, our MORF enables different trees to have specific weights through the guidance of the meta data. 

\subsection{Meta Learning}
Meta-learning is tailored for learning the meta knowledge from the predefined meta task set or meta dataset. It is widely used in few-shot learning. Model-agnostic meta-learning (MAML)~\cite{maml} learns the parameter initialization from the few-shot tasks, and the new tasks only take a few training steps while achieving better generalization than fine-tuning. Jamal~\etal proposed extending the MAML to avoid overfitting on existing training tasks by proposing a maximum-entropy prior that introduces some inequality measures in the loss function~\cite{jamal2019task}. Liu~\etal enhances the generalizability of the main task from the predefined auxiliary task using meta-learning\cite{liu2019self}. Meta-weight-net is a novel weighting network to address class imbalance and noisy label problems~\cite{shu2019meta}. In summary, these meta-learning methods involve optimizing two groups of parameters jointly. The meta training algorithm of our MORF is similar to that of~\cite{shu2019meta}, but differs in its construction of the meta data and weighting behavior over the decision trees. The construction of the meta data in~\cite{shu2019meta} is to select a subset of the validation set with an equal number of samples for each class for meta training, whereas the meta data in MORF is feature level.

\section{Methods}
\label{sec:methods}
In this section, we first formulate the problem of ordinal regression forest for medical images. Then we introduce the meta ordinal regression forest (MORF) framework in order of the training objective, TWW-Net, and GFS module. Finally, we present the meta-learning optimization algorithm and the corresponding theoretical analysis.

\subsection{Problem Formulation}
Ordinal regression solves the problem that the data belonging to different classes have an ordered label, which implies that an intrinsic ordinal relationship exists among the data. It learns a mapping function $h:\mathcal{X} \to \mathcal{Y}$, where $\mathcal{X}$ represents the input space and $\mathcal{Y}$ is the ordinal label space. Here, $\mathcal{Y} = \{y_{1}, y_{2}, \ldots, y_{C}\}$ has the ordinal relationship $y_{1} \preceq y_{2} \preceq \ldots \preceq y_{C}$, where $C$ is the number of  classes. In this study, $y_{c}\in\mathcal{Y}$ denotes the stage of the progression of diseases; taking the lung nodule classification~\cite{LIDC} as an example, $C$ equals $3$ and $y_{1}$, $y_{2}$, and $y_{3}$ represent benign, unsure and malignant, respectively. 

To solve this kind of ordinal classification problem, the given label $y\in\mathcal{Y}$ can be converted to an ordinal distribution label, \ie, a one-dimensional vector $\vct{d} = (d^{1}, d^{2}, \ldots, d^{C-1})\T \in \mathcal{D}$~\cite{zhu2021convolutional}, where $d^{c} = 1$ if $y>y_{c}$, otherwise $d^{c} = 0$. Practically, we will obtain an accurate $\vct{d}$ for a given image, and the $\vct{d}$ should maintain a monotonically decreasing property across all the elements. Therefore, we impose a constraint $d^{1}\ge d^{2} \ge \ldots \ge d^{C-1}$ on $\vct{d}$ during training~\cite{zhu2021convolutional}. Under the framework of the ordinal regression forest (ORF)~\cite{shen2018deep,shen2017label,zhu2021convolutional}, the ordinal label $\vct{d}_{c}$ is given by the leaf nodes in the ORF, and the probability of the given sample $\vct{x}$ falling into the $l$-th leaf node is defined as: 
\begin{equation}
p(l | \vct{x}; \mat{\theta}) = \prod_{n \in \mathcal{N}} s_{n}(\vct{x}; \mat{\theta})^{\vct{1}[l \in \mathcal{L}_{n}^{l}]} \left(1 -  s_{n}(\vct{x}; \mat{\theta})^{\vct{1}[l \in \mathcal{L}_{n}^{r}]}\right),
\label{falling}
\end{equation}
where $\mathcal{L}_{n}^{l}$ and $\mathcal{L}_{n}^{r}$ represent the subsets of leaf nodes held by left and right subtrees of the split node $n$, respectively, and $\mathcal{N}$ is the number of split nodes in one tree. $s_{n}$ is the split function that determines which node (left or right) a sample should be assigned to, \ie, $s_{n}:\mathcal{X} \to \{0, 1\}$. Following~\cite{shen2018deep,kontschieder2015deep,shen2017label,zhu2021convolutional}, we formulate $s_{n}$ as a probabilistic function: $s_{n}(\vct{x};\mat{\theta}) = \sigma(f_{\eta(n)}(\vct{x};\mat{\theta}))$, where $\sigma(\cdot)$ is the sigmoid function, $f(\cdot; \mat{\theta})$ is the backbone network (\eg, a CNN with a FC layer on the top) and the output of $f(\cdot; \mat{\theta})$ is a one-dimensional vector; $\eta(n)$ is an index function that is used to assign elements selected from this one-dimensional vector to all split nodes in the forest. Actually, $\eta(n)$ is implemented as randomly selecting an $\eta(n)$-th element for each $n$-th split node.

When we obtain the probability $p(l | \vct{x}; \mat{\theta})$, the output of one tree can be defined as a mapping $\vct{g} : \mathcal{X} \to \mathcal{D}$:
\begin{equation}
\vct{g}(\vct{x}; \mat{\theta}, \mat{\mathcal{T}}) = \sum_{l \in \mathcal{L}} p(l | \vct{x}; \mat{\theta}) \vct{\pi}_{l},
\label{orf}
\end{equation}
where $\mat{\mathcal{T}}$ denotes one decision tree, and $\mathcal{L}$ a set of leaf nodes. $\vct{\pi}_{l}$ holds the ordinal distribution of the $l$-th leaf node; \ie, $\vct{\pi}_l = (\pi_{l}^{1}, \pi_{l}^{2}, \ldots, \pi_{l}^{C-1})\T$. In this paper, the parameter $\vct{\pi}$ can be updated jointly with that of the backbone network through back-propagation, which has been illustrated in ~\cite{shen2018deep,shen2017label,zhu2021convolutional}. Then, the final prediction of the forest is the average of outputs of all trees:
\begin{equation}
P(\vct{x}) = \frac{1}{T} \sum_{t = 1}^{T} \vct{g}(\vct{x}; \mat{\theta}, \mat{\mathcal{T}}_t),
\end{equation}
where $T$ is the total number of trees. Here, all the trees contribute equally to the final prediction as well as that in previous deep forest methods such as the CORF~\cite{zhu2021convolutional}, whereas in our MORF model, we assign each tree a specific learned weight.

\begin{figure*}
\centering
      \includegraphics[width = 0.8\linewidth]{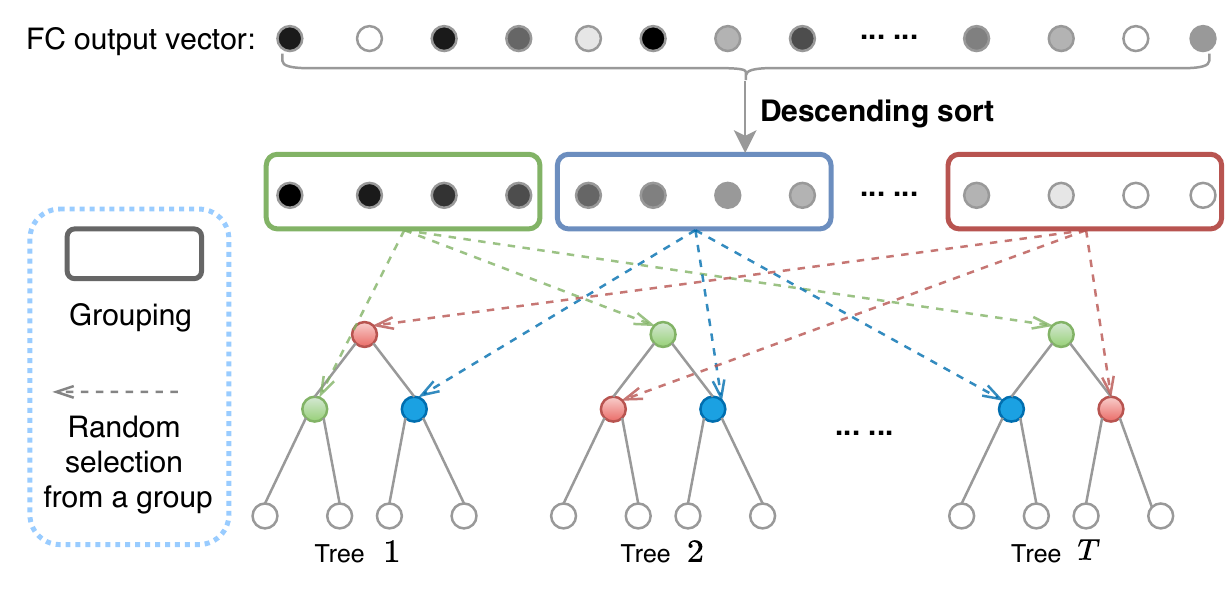}
\caption{The proposed grouped feature selection (GFS) assigning each tree with features of different groups. Note that the number of groups (colored boxes) in GFS equals to the number of split nodes in one decision tree.}
\label{fig:select}
\end{figure*}

\subsection{Meta Ordinal Regression Forests}

The total framework of the MORF contains a CNN with an FC layer as backbone network parameterized by $\vct{\theta}$, a TWW-Net parameterized by $\vct{\phi}$, and the leaf nodes parameterized by $\vct{\pi}$. Note that the $\vct{\pi}$ is updated according to~\cite{zhu2021convolutional}. 

\subsubsection{Objective function}
As mentioned above, all trees in ORF and CORF are assigned the same weights, which can increase the inevitable tree-wise prediction variance. 
To cope with this drawback, we propose multiplying the tree-wise losses with specific weights. Therefore, the gradients of $\vct{\theta}$, backpropagated from the losses of all the trees, can be affected by the weights $\omega_{t}$. Therefore, our training objective function is defined as:
\begin{align}
\vct{\theta}^{*} = \argmin_{\vct{\theta}} L^{\mathrm{tr}}(\vct{\theta}) = \frac{1}{N}\sum_{i = 1}^{N} \frac{1}{T}\sum_{t = 1}^{T} \omega_{t}^{i}  R_{t}^{i}(\vct{\theta}),
\label{eq:obj_main}
\end{align}
where $N$ is the number of training images, $R_{t}^{i}$ denotes the classification loss generated by~\eqref{orf}, and  $\omega_{t}$ represents the specific weight for the $t$-th tree that is  learned by the TWW-Net, which will be subsequently introduced. Different from~\cite{shu2019meta}, equation~\eqref{eq:obj_main} imposes the weights on the different trees w.r.t. one training sample $i$, rather than on the different samples.

\subsubsection{Tree-wise weighting network}
Here, we introduce the TWW-Net that is used for learning the weights $\omega_{t}$ in~\eqref{eq:obj_main}. Similar to the meta-weight-net~\cite{shu2019meta}, TWW-Net is practically implemented as a group of multilayer perceptrons (MLPs), $V_{t}$, and each MLP acts as a weight learning function for a specific tree because of the universal approximation property of MLPs~\cite{csaji2001approximation}. In Fig.~\ref{fig:framework}, we can see that the $t$-th tree generates a classification loss $R_{t}$, then $R_{t}$ will be fed into the corresponding weighting net $V_{t}$, and finally $V_{t}$ outputs the weight $w_{t}$ for the $t$-th tree. This process can be formulated as:
\begin{align}
w_{t} = V_{t} (R_{t} (\vct{\theta}) ; \vct{\phi}_{t}).
\label{eq:w_t}
\end{align}
Therefore, TWW-Net is composed of $T$ weighting net $V_{t}$, where $T$ is the number of trees. Here we use $\vct{\phi} = \{ \vct{\phi}_{1}, \vct{\phi}_{2}, \ldots, \vct{\phi}_{T} \}$ to represent the set of parameters of TWW-Net, and they are updated together which will be described in Section~\ref{subsec:optimization}. Through~\eqref{eq:w_t}, a TWW-Net can assign different weights to different trees.

Combined with~\eqref{eq:obj_main}, the training  objective function can be modified as follows:
\begin{align}
\label{eq:theta_obj}
 \vct{\theta}^{*}(\vct{\phi}) &= \argmin_{\vct{\theta}} L^{\mathrm{tr}}(\vct{\theta}; \vct{\phi}) \\
 &= \frac{1}{N}\sum_{i=1}^{N} \frac{1}{T}\sum_{t=1}^{T} V_{t}^{i}\left[R_{t}^{i}(\vct{\theta}, \vct{\pi}; S^{\mathrm{tr}}); \vct{\phi}_{t}\right] \cdot R_{t}^{i}(\vct{\theta}, \vct{\pi}; S^{\mathrm{tr}}),\notag
\end{align}
where $S^{\mathrm{tr}}$ denotes training set.

\subsubsection{Grouped feature selection (GFS)}
Although we have incorporated the tree-wise weights in the training objective function, the structure of the forest is still fixed. Therefore, we introduce the GFS module to construct the dynamic forest with random feature perturbation in this section. Then, we explore the relationship between the GFS and random forest.

As shown in Fig.~\ref{fig:select}, the GFS first ranks all the activation values of the final FC feature vector. Then it splits the ranked elements into $\mathcal{N}$ groups (denoted by different colors), where $\mathcal{N}$ equals the number of split nodes in one decision tree. Both the elements inside and outside of one group are in descending order, and each tree randomly selects its own nodes across all the groups. Hence, one tree contains the features globally across the FC feature, and it retains the local random perturbation of the feature that is critical for the random forest. After repeating these procedures on all the trees,  the dynamic forest is constructed. Note that the final trained model of MORF is also equipped with a forest that is fixed from the beginning of training, and the GFS module only works in the training stage and has no impact on the forest at the time of inference. Note that, the number of groups corresponds to the number of split nodes in one tree, \ie, the number of groups increases along with an increase of tree depth.

\noindent\textbf{Relationship between GFS and Random Forest.}\quad  Random forest (RF) is a classical ensemble learning method, which benefits from base learners that have feature and data random perturbations. Specifically, feature perturbation means that each node in the decision trees is the most discriminative attribute in a subset of its whole attribute set~\cite{breiman2001random}. Data perturbation is satisfied for all the deep forest-based methods that trained over shuffled mini-batches, however, feature perturbation occurs because of the fixed forest structure~\cite{shen2018deep,shen2017label,zhu2021convolutional}. In Fig.~\ref{fig:select} \textbf{(b)}, we can see that each split node is randomly selected from its own feature set. Although, in Fig.~\ref{fig:select} \textbf{(a)}, all the nodes in the forest are also obtained through random selection within their subsets (indicated by different colors) of the FC feature, it differs from the RF in that the nodes in different trees share the same subsets, \ie, the GFS-based forest also maintains the node-wise feature random perturbation. Therefore, the MORF with GFS possesses the merit of randomness with respect to all the split nodes, and this advantage does not exist for previous methods~\cite{shen2018deep,kontschieder2015deep,shen2017label,zhu2021convolutional}.

\subsection{Optimization via Meta-learning}
\label{subsec:optimization}
When we obtain a dynamic forest, we expect it to guide the update of the CNN with a fixed forest. Moreover, from~\eqref{eq:theta_obj} we observe that the objective function involves two parts of parameters, $\vct{\theta}$ and $\vct{\phi}$, and $\vct{\theta}$ is a function of $\vct{\phi}$, so we customize a meta-learning framework  enabling the meta data to guide the learning of the target model. Here, the GFS selected features are regarded as the meta data, which is different from those in~\cite{shu2019meta}.

To obtain the optimal $\vct{\theta}^{*}$, we need to obtain the optimal $\vct{\phi}^{*}$. Therefore, we optimize $\vct{\phi}$ by minimizing the following objective function:
\begin{align}
\label{eq:phi_obj}
 \vct{\phi}^{*} &= \argmin_{\vct{\phi}} L^{\mathrm{meta}}(\vct{\theta}^{*}(\vct{\phi})) \\
 &= \frac{1}{M} \sum_{j=1}^{M} \frac{1}{T} \sum_{t=1}^{T} R_{t}^{j}(\vct{\theta}^{*}(\vct{\phi}_{t}), \vct{\pi}; S^{\mathrm{meta}}),\notag
\end{align}
where $M$ is the number of meta data. This objective function indicates that $\vct{\phi}$ is updated based on the optimal backbone parameter $\vct{\theta}^{*}$. 

First, we take the derivative of~\eqref{eq:theta_obj} with respect to $\vct{\theta}$: 
\begin{align}
 \vct{\widehat{\theta}}^{(u)} &= \vct{\theta}^{(u)}- \label{eq:theta_hat}\\ 
 & \alpha \frac{1}{N \cdot T} \sum_{i=1}^{N} \sum_{t=1}^{T} V_{t}^{i}\left[R_{t}^{i}(\vct{\theta}^{(u)}); \vct{\phi}_{t}\right] \cdot \nabla_{\vct{\theta}} R_{t}^{i}(\vct{\theta}) \Big\rvert_{\vct{\theta}^{(u)}},\notag
\end{align}
where $\alpha$ is the learning rate for $\vct{\theta}$. For simplicity, we omit the parameters $\vct{\pi}$ and the datasets $S^{\mathrm{tr}}$ and $S^{\mathrm{meta}}$ in the above equations. The superscript $(u)$ denotes the $u$-th iteration. Therefore, $\widehat{\vct{\theta}}$ in~\eqref{eq:theta_hat} represents the weights obtained through the first order derivative $\nabla_{\vct{\theta}} R_{t}^{i}(\vct{\theta})$. Then we can use $\widehat{\vct{\theta}}$ to update the parameters $\vct{\phi}$:
\begin{align}
\vct{\phi}^{(u+1)} = \vct{\phi}^{(u)} - \beta \frac{1}{M} \sum_{j=1}^{M} \nabla_{\vct{\phi}}\left[\frac{1}{T} \sum_{t=1}^{T} R_{t}^{j}(\widehat{\vct{\theta}}^{(u)})\right] \Big|_{\vct{\phi}^{(u)}}.
\label{eq:update_phi}
\end{align}

To go a step further, we derive~\eqref{eq:update_phi} and obtain the following equation:
\begin{align}
\vct{\phi}^{(u+1)}& = \vct{\phi}^{(u)} + \label{eq:new_update_phi}\\
&\frac{\alpha\beta}{N} \sum_{i=1}^{N} \left(\frac{1}{M} \sum_{j=1}^{M} G_{ij}\right) \frac{\sum_{t=1}^{T}\partial{V_{t}^{i}(R_{t}^{i}(\vct{\theta}^{(u)};\vct{\phi}_{t}))}}{\partial{\vct{\phi}}} \Big|_{\vct{\phi}^{(u)}},\notag
\end{align}
where 
\begin{align*}
G_{ij} = \frac{1}{T}\sum_{t=1}^{T} \frac{\partial{R_{t}^{j}(\widehat{\vct{\theta}}, \vct{\pi}; S^{\mathrm{meta}})}}{\partial{\widehat{\vct{\theta}}}} |_{\widehat{\vct{\theta}}^{(u)}} \cdot \frac{1}{T}\sum_{t=1}^{T} \frac{\partial{R_{t}^{i}(\vct{\theta}; S^{\mathrm{tr}})}}{\partial{\vct{\theta}}}|_{\vct{\theta}^{(u)}}
\end{align*}
 stands for the similarity between two gradients---the gradient of the $i$-th training data computed on training loss $R_{t}^{i}$ and the gradient of the mean value of the mini-batch meta data calculated on meta loss $R_{t}^{j}$. This enforces the gradient of the feature of training data to approach that of meta data generated from GFS. Hence, the behavior of each tree is guided by the meta gradient and is consistent with other trees. Consequently, the predictions of different trees in our MORF are consistent, \ie, have lower variance, which guarantees a more stable prediction.

After we obtain the updated TWW-Net parameters $\vct{\phi}^{(u+1)}$, the update rule of $\vct{\theta}$ can be defined as:
\begin{align}
\label{eq:update_theta}
 \vct{\theta}^{(u+1)} &= \vct{\theta}^{(u)} - \\
 &\alpha \frac{1}{NT} \sum_{i=1}^{N}\sum_{t=1}^{T} V_{t}^{i}\left[R_{t}^{i}(\vct{\theta}^{(u)}); \vct{\phi}^{(u+1)}_{t}\right] \nabla_{\vct{\theta}} R_{t}^{i}(\vct{\theta}^{(u)}) \Big|_{\vct{\theta}^{(u)}}.\notag
\end{align}

\begin{algorithm}[t]

\caption{Training Algorithm of MORF.}
\label{alg:train}
\LinesNumbered
\KwIn{Training data $S^{\mathrm{tr}}$, max iteration $U$.}
\KwOut{Ordinal regressor parameter $\vct{\theta}^{(U)}$.}
Initialize  $\vct{\theta}^{(0)}$ and  $\vct{\phi}^{(0)}$.\\
\For{$u=0$ to $U-1$}{
Sample a mini-batch of ($\vct{x}$, $\vct{d}$) from $S^{\mathrm{tr}}$.\\
\emph{// Meta train (update $\mat{\phi}$):}\\
Forward: input ($\vct{x}$, $\vct{d}$) with $\vct{\theta}^{(u)}(\vct{\phi}^{(u)})$.\\
Randomly construct the forest $F^{R}$ and compute tree-wise loss $R_{t}(\vct{\theta}^{(u)}(\vct{\phi}^{(u)});S^{\mathrm{tr}})$.\\
Compute the first order derivative $\widehat{\vct{\theta}}^{(u)}(\vct{\phi}^{(u)})$ by~\eqref{eq:theta_hat}.\\
Forward: input ($\vct{x}$, $\vct{d}$) with $\widehat{\vct{\theta}}^{(u)}(\vct{\phi}^{(u)})$.\\
Construct the forest via GFS and compute tree-wise loss $R_{t}(\widehat{\vct{\theta}}^{(u)}(\vct{\phi}^{(u)});S^{\mathrm{meta}})$.\\
Update $\vct{\phi}^{(u+1)}$ by~\eqref{eq:update_phi}.\\
\emph{// Model train (update $\mat{\theta}$):}\\
Forward: input ($\vct{x}$, $\vct{d}$) with $\vct{\theta}^{(u)}(\vct{\phi}^{(u+1)})$.\\
Compute tree-wise loss $R_{t}(\vct{\theta}^{(u)}(\vct{\phi}^{(u+1)});S^{\mathrm{tr}})$ based on the constructed $F^{R}$ in Step \textbf{6}.\\
Update $\vct{\theta}^{(u+1)}$ by~\eqref{eq:update_theta}.\\
}
\end{algorithm}
In~\eqref{eq:update_phi}, the tree-wise loss $R_{t}(\widehat{\vct{\theta}}^{(u)}, \vct{\pi}; S^{\mathrm{meta}})$ is calculated via meta data, and the meta data are feature level. During the training procedure, we first obtain the first-order derivative $\widehat{\vct{\theta}}(\vct{\phi})$ in~\eqref{eq:theta_hat} via taking the image $\vct{x}$ as input, and the forest is constructed based on this forward process as shown on the left side of Fig.~\ref{fig:framework}. Then we fix $\widehat{\vct{\theta}}(\vct{\phi})$ in~\eqref{eq:theta_hat} and take the image $\vct{x}$ as input, and the forest here is reconstructed through our GFS module as shown on the right side of Fig.~\ref{fig:framework}. Once again, the globally and locally selected features via GFS are the meta data in our method. This retains the structural variability and the random feature perturbation of the forest. Simultaneously, our training scheme can guide the behavior of learning from training data to approach that of learning from the GFS generated meta data. The details of the training procedure are shown in Algorithm~\ref{alg:train}.

Note that the TWW-Net ($\vct{\phi}$) only works in the meta training stage, and it does not affect the prediction in inference. In other words, the trained model contains parameters $\vct{\theta}$ and $\vct{\pi}$, which are required during inference.


\begin{table*}[t]

\caption{Classification results on test set of Train(3)-Test(3) on LIDC-IDRI dataset. The values with underlines indicate the best results while less important in the clinical diagnosis~\cite{wu2019learning}. MORF$_c$ is the conference version of the proposed MORF~\cite{lei2020meta}.}
\centering
\scriptsize
\begin{tabular*}{1.0\linewidth}{@{\extracolsep{\fill}}cl*{10}{c}}
\toprule
\multirow{2}{*}{Backbone} & \multirow{2}{*}{Method} & \multirow{2}{*}{Accuracy} & \multicolumn{3}{c}{Benign} & \multicolumn{3}{c}{Malignant} & \multicolumn{3}{c}{Unsure} \\ \cline{4-12}
 &  &  & Precision & Recall & F1 & Precision & Recall & F1 & Precision & Recall & F1 \\
\midrule
\multirow{6}*{\makecell{ResNet-18}}
& CE Loss & 0.542$\pm0.006$ & 0.544 & 0.722 & 0.620 & 0.586 & 0.644 & 0.613 & 0.496 &  0.290 & 0.366 \\
& Poisson~\cite{beckham2017unimodal} & 0.527$\pm0.007$ & 0.536 & 0.605 & 0.568 & 0.590 & 0.584 & 0.587 & 0.477 & 0.410 & 0.441 \\
& NSB~\cite{liu2018ordinal} & 0.534$\pm0.007$ & 0.517 & \underline{0.807} & 0.630 & 0.607 & 0.673 & 0.638 & 0.500 & 0.160 & 0.242 \\
& UDM~\cite{wu2019learning} & 0.546$\pm0.004$ & 0.553 & 0.767 & 0.643 & 0.581 & 0.495 & 0.535 & 0.504 & 0.325 & 0.395 \\
& CORF~\cite{zhu2021convolutional} & 0.568$\pm0.004$ & 0.569 & 0.713 & 0.633 & 0.632 & 0.613 & 0.623 & \underline{0.523} & 0.385 & 0.443 \\
& MORF$_c$~\cite{lei2020meta} & 0.573$\pm0.004$ & 0.695 & 0.623 & 0.657 & 0.627 & 0.683 & 0.654 & 0.479 & \textbf{0.685} & \textbf{0.564} \\
& \textbf{MORF (ours)} & \textbf{0.634$\pm0.003$} & \textbf{0.706} & 0.691 & \textbf{0.698} & \underline{0.783} & \textbf{0.752} & \textbf{0.768} & 0.412 & 0.441 & 0.426 \\
\midrule
\multirow{7}*{\makecell{ResNet-34}}
& CE Loss & 0.534$\pm0.007$ & 0.540 & \underline{0.722} & 0.618 & 0.625 & 0.644 & 0.634 & 0.443 & 0.270 & 0.335 \\
& Poisson~\cite{beckham2017unimodal} & 0.540$\pm0.006$ & 0.577 & 0.502 & 0.537 & 0.714 & 0.545 & 0.618 & 0.458 & 0.520 & 0.512 \\
& NSB~\cite{liu2018ordinal} & 0.544$\pm0.006$ & 0.580 & 0.619 & 0.599 & 0.608 & 0.614 & 0.611 & 0.462 & 0.425 & 0.443 \\
& UDM~\cite{wu2019learning} & 0.536$\pm0.004$ & 0.537 & 0.659 & 0.592 & 0.578 & 0.554 & 0.566 & \underline{0.510} &  0.390 & 0.442 \\
& CORF~\cite{zhu2021convolutional} & 0.542$\pm0.005$ & 0.581 & 0.605 & 0.593 & 0.616 & 0.524 & 0.566 & 0.466 & 0.480 & 0.473 \\
& MORF$_c$~\cite{lei2020meta} & 0.586$\pm0.004$ & 0.619 & 0.619 & 0.619 & 0.504 & 0.700 & 0.565 & 0.509 & \textbf{0.560} & \textbf{0.533} \\
& \textbf{MORF (ours)} & \textbf{0.639$\pm0.003$} & \textbf{0.751} & 0.650 & \textbf{0.697} & \underline{0.773} & \textbf{0.742} & \textbf{0.757} & 0.422 & 0.535 & 0.472 \\
\midrule
\multirow{7}*{\makecell{VGG-16}}
& CE Loss & 0.517$\pm0.008$ & 0.538 & 0.668 & 0.596 & 0.562 & 0.495 & 0.526 & 0.456 & 0.360 & 0.402 \\
& Poisson~\cite{beckham2017unimodal} & 0.542$\pm0.005$ & 0.548 & \underline{0.794} & 0.648 & 0.568 & 0.624 & 0.594 & 0.489 & 0.220 & 0.303 \\
& NSB~\cite{liu2018ordinal} & 0.553$\pm0.005$ & 0.565 & 0.641 & 0.601 & 0.566 & 0.594 & 0.580 & \underline{0.527} & 0.435 & 0.476 \\
& UDM~\cite{wu2019learning} & 0.548$\pm0.002$ & 0.541 & 0.767 & 0.635 & 0.712 & 0.515 & 0.598 & 0.474 & 0.320 & 0.382 \\
& CORF~\cite{zhu2021convolutional} & 0.559$\pm0.006$ & 0.590 & 0.627 & 0.608 & 0.704 & 0.495 & 0.581 & 0.476 & 0.515 & 0.495 \\
& MORF$_c$~\cite{lei2020meta} & 0.580$\pm0.004$ & 0.660 & 0.479 & 0.556 & 0.719 & 0.596 & 0.652 & 0.492 & \textbf{0.680} & \textbf{0.571} \\
& \textbf{MORF (ours)} & \textbf{0.641$\pm0.002$} & \textbf{0.754} & 0.632 & \textbf{0.688} & \underline{0.798} & \textbf{0.743} & \textbf{0.769} & 0.429 & 0.574 & 0.492 \\
\bottomrule
\end{tabular*}
\label{tab:train3_test3}
\end{table*}

\section{Experimental Setup and Results}
\label{sec:experiments}
In this section, we present the experimental setup and evaluate the proposed MORF method on the LIDC-IDRI~\cite{LIDC} and BUSI~\cite{al2020dataset} datasets. The experiments reported the average results through five randomly independent  split folds, and for the accuracies, we reported the average values with standard deviations.

\begin{figure}[t]
\centering
\includegraphics[width = 1.0\linewidth]{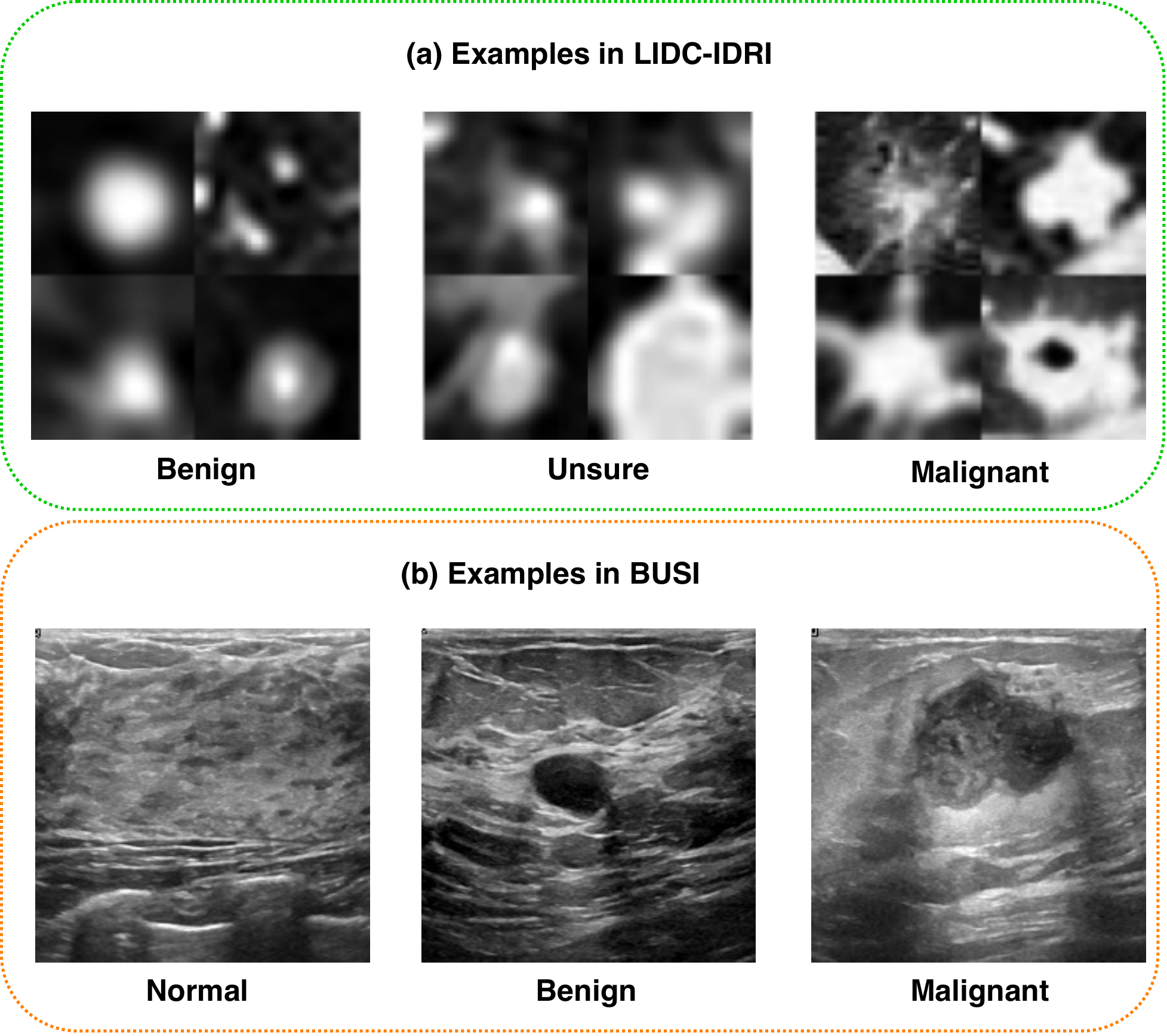}
\caption{Some examples in the BUSI~\cite{al2020dataset} and LIDC-IDRI~\cite{LIDC} datasets. For the LIDC-IDRI dataset, we provide four examples each class.}
\label{fig:examples}
\end{figure}

\subsection{Data Preparation}
\label{subsec:data}
\subsubsection{LIDC-IDRI dataset for lung nodule classification}
LIDC-IDRI is a publicly available dataset for pulmonary nodule classification or detection, which involves 1,010 patients. Some representative cases are shown in Fig.~\ref{fig:examples}. All the nodules were labeled by four radiologists, and each nodule was rated with a score from 1 to 5, indicating malignant progression of the nodule. In our experiments, we cropped the ROI with a square shape of a doubled equivalent diameter at the corresponding center of a nodule. The averaged score of a nodule was used as the ground-truth label during training. Note that the averaged scores also range from 1 to 5, and we regard a nodule with an average score between 2.5 and 3.5 as the unsure nodule, benign and malignant nodules are those with scores that are lower than 2.5 and higher than 3.5, respectively~\cite{wu2019learning}.  In each plane, all the CT volumes were preprocessed to have $1mm$ spacing in each plane. Finally, we obtain the training and testing data by cropping the $32 \times 32 \times 32$ volume ROIs located at the annotated centers.

\subsubsection{BUSI dataset for breast cancer classification}
The BUSI dataset can be used for ultrasound image-based breast cancer classification and segmentation, which contains 780 images of three classes: 133 normal, 487 benign, and 210 malignant images. Some representative cases are shown in Fig.~\ref{fig:examples}. We first resized the original 2-$D$ images into the same sizes $128 \times 128$, and then conducted the data augmentations, including flipping and adding random Gaussian noise, for the training set of BUSI. Finally, the training and test sets contain 1,872 images and 156 images for each fold, respectively.

\subsection{Implementation Details}
\label{subsec:implement}
\subsubsection{Network architecture} We applied ResNet-18, ResNet-34, and VGG-16~\cite{ResNet,vgg} as backbone networks to compare our MORF with other methods. Because the scales of the two datasets are relatively small, we use the 2-$D$ version of the backbone networks to avoid the huge number of parameters in 3-$D$ networks. Therefore, the input of the model, the $32 \times 32 \times 32$ volumes, can be treated as $32 \times 32$ patches with $32$ channels each, and the corresponding number of channels of the first layer is set as $32$. For our MORF, the output dimension of the final FC layer equals to the number of split nodes in a forest due to the GFS using random selection without replacement, and for CORF, it is set as $256$.

\begin{table*}[t]
\scriptsize
\caption{Classification results on test sets of Train(3)-Test(2) and Train(2)-Test(2) on LIDC-IDRI dataset. The values with underlines indicate the best results while less important in the clinical diagnosis~\cite{wu2019learning}. In this Table, P., R., and F1 are abbreviations of Precision, Recall, and F1 score, respectively. B. is short for Backbone. MORF$_c$ is the conference version of the proposed MORF~\cite{lei2020meta}.}
\centering
\begin{tabular*}{1.0\textwidth}{@{\extracolsep{\fill}}cl*{7}{c}*{7}{c}}
\toprule
\multicolumn{2}{c}{}&\multicolumn{7}{c}{Train(3)-Test(2)} & \multicolumn{7}{c}{Train(2)-Test(2)} \\ \hline
\multirow{2}{*}{B.} & \multirow{2}{*}{Method} & \multirow{2}{*}{Acc.} & \multicolumn{3}{c}{Benign} & \multicolumn{3}{c}{Malignant} & \multirow{2}{*}{Acc.} & \multicolumn{3}{c}{Benign} & \multicolumn{3}{c}{Malignant}\\ \cline{4-9}\cline{11-16}
 &  &  & P. & R. & F1  & P. & R. & F1  &  & P. & R. & F1  & P. & R. & F1  \\
\midrule
\multirow{6}*{\makecell{\rotatebox{90}{ResNet-18}}}
& CE Loss & 0.833$\pm0.006$ & 0.848 & 0.924 & 0.884 & 0.790 & 0.634 & 0.703 & 0.855$\pm0.005$ & 0.879 & 0.915 & 0.897 & 0.793 & 0.723 & 0.756\\
& Poisson~\cite{beckham2017unimodal} & 0.818$\pm0.004$ & 0.820 & 0.942 & 0.877 & 0.809 & 0.545 & 0.651 & 0.840$\pm0.003$ & 0.887 & 0.879 & 0.883 & 0.738 & 0.752 & 0.745\\
& NSB~\cite{liu2018ordinal} & 0.781$\pm0.005$ & 0.814 & 0.906 & 0.858 & 0.761 & 0.505 & 0.607 & 0.849$\pm0.007$ & 0.875 & 0.910 & 0.892 & 0.783 & 0.713 & 0.746\\
& UDM~\cite{wu2019learning} & 0.793$\pm0.004$ & 0.855 & 0.870 & 0.862 & 0.741 & 0.624 & 0.677 & 0.846$\pm0.005$ & 0.888 & 0.888 & 0.888 & 0.752 & 0.752 & 0.752 \\
& CORF~\cite{zhu2021convolutional} & 0.815$\pm0.003$ & 0.811 & \underline{0.959} & 0.878 & \underline{0.862} & 0.495 & 0.628 & 0.830$\pm0.002$ & 0.875 & 0.879 & 0.877 & 0.730 & 0.723 & 0.726\\
& MORF$_c$~\cite{lei2020meta} & 0.839$\pm0.004$ & 0.861 & 0.945 & 0.886 & 0.781 & 0.673 & 0.723 & 0.848$\pm0.005$ & 0.881 & 0.901 & 0.891 & 0.771 & 0.733 & 0.751 \\
& \textbf{MORF (Ours)} & \textbf{0.889$\pm0.004$} & \textbf{0.923} & 0.915 & \textbf{0.919} & 0.816 & \textbf{0.832} & \textbf{0.824} & \textbf{0.886$\pm0.003$} & \textbf{0.915} & \underline{0.919} & \textbf{0.917} & \underline{0.820} & \textbf{0.812} & \textbf{0.816} \\
\midrule
\multirow{6}*{\makecell{\rotatebox{90}{ResNet-34}}}
& CE Loss & 0.824$\pm0.005$ & 0.852 & 0.928 & 0.888 & 0.811 & 0.594 & 0.686 & 0.846$\pm0.005$ & 0.862 & 0.923 & 0.891 & 0.800 & 0.673 & 0.731\\
& Poisson~\cite{beckham2017unimodal} & 0.833$\pm0.004$ & 0.839 & \underline{0.937} & 0.886 & 0.836 & 0.604 & 0.701 & 0.840$\pm0.006$ & 0.867 & 0.906 & 0.886 & 0.769 & 0.693 & 0.729\\
& NSB~\cite{liu2018ordinal} & 0.763$\pm0.006$ & 0.883 & 0.812 & 0.846 & 0.815 & 0.653 & 0.725 & 0.846$\pm0.005$ & 0.856 & 0.933 & 0.893 & 0.815 & 0.653 & 0.725\\
& UDM~\cite{wu2019learning} & 0.821$\pm0.004$ & 0.860 & 0.906 & 0.882 & 0.790 & 0.634 & 0.703 & 0.870$\pm0.003$ & 0.891 & 0.923 & 0.907 & 0.817 & 0.752 & 0.784\\
& CORF~\cite{zhu2021convolutional} & 0.815$\pm0.004$ & 0.867 & 0.883 & 0.875 & 0.752 & 0.663 & 0.705 & 0.851$\pm0.005$ & 0.895 & 0.887 & 0.891 & 0.757 & 0.772 & 0.764\\
& MORF$_c$~\cite{lei2020meta} & 0.830$\pm0.003$ & 0.872 & 0.883 & 0.877 & 0.735 & 0.713 & 0.723 & 0.858$\pm0.004$ & 0.886 & 0.910 & 0.898 & 0.789 & 0.743 & 0.765 \\
& \textbf{MORF (ours)} & \textbf{0.907$\pm0.003$} & \textbf{0.941} & 0.924 & \textbf{0.932} & \underline{0.838} & \textbf{0.871} & \textbf{0.854} & \textbf{0.901$\pm0.003$} & \textbf{0.906} & \underline{0.955} & \textbf{0.930} & \underline{0.887} & \textbf{0.782} & \textbf{0.831} \\
\midrule
\multirow{6}*{\makecell{\rotatebox{90}{VG-16}}}
& CE Loss & 0.775$\pm0.007$ & 0.848 & 0.848 & 0.848 & 0.816 & 0.614 & 0.701 & 0.855$\pm0.005$ & 0.864 & \underline{0.937} & 0.899 & 0.829 & 0.673 & 0.743 \\
& Poisson~\cite{beckham2017unimodal} & 0.784$\pm0.006$ & 0.888 & 0.785 & 0.833 & 0.622 & \textbf{0.782} & 0.693 & 0.824$\pm0.005$ & 0.858 & 0.897 & 0.877 & 0.770 & 0.663 & 0.713\\
& NSB~\cite{liu2018ordinal} & 0.772$\pm0.004$ & 0.839 & 0.917 & 0.876 & 0.863 & 0.624 & 0.724 & 0.852$\pm0.005$ & 0.869 & 0.924 & 0.896 & 0.805 & 0.693 & 0.745\\
& UDM~\cite{wu2019learning} & 0.796$\pm0.003$ & 0.859 & 0.928 & 0.892 & \underline{0.895} & 0.505 & 0.646 & 0.867$\pm0.004$ & 0.898 & 0.910 & 0.904 & 0.796 & 0.772 & 0.784\\
& CORF~\cite{zhu2021convolutional} & 0.796$\pm0.005$ & 0.887 & 0.883 & 0.885 & 0.772 & 0.603 & 0.678 & 0.864$\pm0.007$ & \textbf{0.918} & 0.879 & 0.898 & 0.786 & \textbf{0.802} & \textbf{0.794} \\
& MORF$_c$~\cite{lei2020meta} & 0.824$\pm0.004$ & 0.856 & 0.892 & 0.875 & 0.739 & 0.673 & 0.705 & 0.870$\pm0.003$ & 0.906 & 0.906 & 0.906 & 0.792 & 0.792 & 0.792 \\
& \textbf{MORF (ours)} & \textbf{0.877$\pm0.004$} & \textbf{0.889} & \underline{0.937} & \textbf{0.912} & 0.842 & 0.742 & \textbf{0.789} & \textbf{0.877$\pm0.004$} & 0.893 & 0.933 & \textbf{0.912} & \underline{0.835} & 0.752 & 0.792 \\
\bottomrule
\end{tabular*}
\label{tab:train32_test2}
\end{table*}

\subsubsection{Hyperparameter setting} The learning rates for the LIDC-IDR and the BUSI datasets are $0.001$ and $0.00005$, respectively, and are decayed by $0.1$ every $120$ epochs (150 epochs in total); the sizes of a mini-batch size are $16$, and the weight decay values for the Adam optimizer are $0.0001$ and $0.00005$, respectively~\cite{kingma2014adam}. The loss functions used in the MORF, CORF, and each tree-wise loss during meta training are the standard CE loss~\cite{crossentropy,zhang2019medical,pesce2019learning}.

The number of trees for the forest is $4$ and the tree depth is $3$. In practice, the TWW-Net contains several MLPs, where the number of MLPs equals the number of trees. For the LIDC-IDRI dataset, we evaluated whether or not the unsure nodules were used for training or testing. Therefore, in Tables~\ref{tab:train3_test3} and~\ref{tab:train32_test2}, we use the symbol Train($n_{1}$)-Test($n_{2}$) to represent that there are $n_{1}$ classes of data for training, and $n_{2}$ classes for testing. The value of $n_{1}$ and $n_{2}$ is $3$ (with unsure data) or $2$ (without unsure data). All of our experiments are implemented with the PyTorch~\cite{pytorch} framework and trained with an NVIDIA GTX 2080 Ti GPU.

\subsection{Training with Unsure Data for Lung Nodule Classification}
In this section, we focus on the standard 3-class classification of lung nodules. Following~\cite{wu2019learning}, we also care more about the recall of malignant lesions and the precision of benign lesions, which fits more appropriately with the clinical diagnosis. 

In Table~\ref{tab:train3_test3}, we illustrate the results of the CE loss-based methods, the related ordinal regression methods, and our MORF as well as its conference version denoted as MORF$_c$. When using different backbones, the MORF achieves the best accuracies and F1 scores of malignant and benign. For  all the backbones, the MORF also maintains the higher recall of malignant and precision of benign. Under the meaning of the clinical diagnosis, the MORF is better able to reduce the missing diagnosis rate, \ie, there will be fewer unsure and malignant nodules diagnosed as benign, and fewer malignant nodules missing a diagnosis. Most importantly, MORF obtains the best precision for malignancy, demonstrating a lower misdiagnosis rate.

Under the setting of Train(3)-Test(2) where training data includes all unsure data, the feature space will be more complicated compared with the binary classification setting. However, the left part of Table~\ref{tab:train32_test2} shows that the MORF significantly outperforms the other methods significantly on all measured metrics. We emphasize that the ordinal relationship of the data is critical to ordinal regression which can be regarded as a fine-grained classification, and the accurate feature representation determines the ability of the final classifier. Although both MORF and CORF~\cite{zhu2021convolutional} consider the global ordinal relationship, the fixed forest of the CORF degrades its performance in that the random feature perturbation is omitted. The MORF with reconstructed forest via the GFS module enable the update of the parameter $\vct{\theta}$ to be affected by the feature randomness, hence leading to a significant gain. Therefore, this experiment verifies the robustness of MORF against the influence of plugging unsure samples into the training data. Most importantly, the results of MORF on the left side of Table~\ref{tab:train32_test2} are slightly better than those on the right side. This indicates that using the unsure class is helpful for improving the classification of the other two classes, \ie, the unsure nodules act as a boundary between the malignant and benign. Especially for ResNet-18/34, the recalls of malignant and the precisions of benign of Train(3)-Test(2) are higher than those of Train(2)-Test(2). VGG-16 achieves comparable results under these two settings, and we attribute this phenomenon to the different feature spaces learned by ResNet and VGG-16.

When comparing the unsure class with existing methods, Table~\ref{tab:train3_test3} shows both MORF and MORF$_c$ achieve better results than other methods in terms of recall rather than precision. Similar to the importance of recall of malignant~\cite{wu2019learning}, the higher recall of the unsure class provides us with significant insights that there will be fewer unsure nodules likely being classified as benign or malignant. 
Although the unsure class contains mixed benign and malignant samples, one should not miss any malignant samples in unsure class. That is, the unsure class should be similar to malignant class that the recall is relatively more important than the precision. 
Therefore, MORF and MORF$_c$ are helpful for further diagnosis of nodules, such as biopsy. Consequently, MORF is more suitable for real clinical circumstance while recommending more accurate diagnosis of follow-ups.
It is noticed that this superiority does not hold in the UDM method~\cite{wu2019learning}. On the other hand, for a certain backbone, both the MORF and MORF$_c$ outperform the CE loss-based counterpart, which exhibits their effectiveness on exploiting ordinal relationship.

\begin{table*}[t]

\caption{Classification results on test set of BUSI dataset. The values with underlines indicate the best results while less important in the clinical diagnosis~\cite{wu2019learning}.MORF$_c$ is the conference version of the proposed MORF~\cite{lei2020meta}.}
\centering
\begin{tabular*}{1.0\textwidth}{@{\extracolsep{\fill}}cl*{10}{c}}
\toprule
\multirow{2}{*}{Backbone} & \multirow{2}{*}{Method} & \multirow{2}{*}{Accuracy} & \multicolumn{3}{c}{Normal} & \multicolumn{3}{c}{Benign} & \multicolumn{3}{c}{Malignant} \\ \cline{4-12}
 &  &  & Precision & Recall & F1 & Precision & Recall & F1 & Precision & Recall & F1\\
\midrule
\multirow{7}*{\makecell{ResNet-18}}
& CE Loss & 0.726$\pm0.003$ & 0.750 & 0.556 & 0.638 & 0.779 & 0.761 & 0.770 & 0.627 & 0.762 & 0.688 \\
& Poisson~\cite{beckham2017unimodal} & 0.707$\pm0.004$ & 0.692 & \underline{0.667} & 0.679 & 0.817 & 0.659 & 0.730 & 0.583 & 0.833 & 0.686 \\
& NSB~\cite{liu2018ordinal} & 0.751$\pm0.005$ & 0.875 & 0.259 & 0.400 & 0.739 & \underline{0.932} & 0.824 & 0.763 & 0.690 & 0.725 \\
& UDM~\cite{wu2019learning} & 0.739$\pm0.003$ & \underline{1.000} & 0.259 & 0.412 & 0.762 & 0.875 & 0.814 & 0.653 & 0.761 & 0.703 \\
& CORF~\cite{zhu2021convolutional} & 0.758$\pm0.006$ & 0.875 & 0.519 & 0.651 & 0.723 & 0.920 & 0.809 & \underline{0.827} & 0.571 & 0.676 \\
& MORF$_c$~\cite{lei2020meta} & 0.764$\pm0.003$ & \underline{1.000} & 0.481 & 0.650 & 0.816 & 0.807 & 0.811 & 0.632 & 0.857 & 0.727 \\
& \textbf{MORF (ours)} & \textbf{0.809$\pm0.003$} & 0.882 & 0.556 & \textbf{0.682} & \textbf{0.850} & 0.841 & \textbf{0.845} & 0.717 & \textbf{0.905} & \textbf{0.800} \\
\midrule
\multirow{7}*{\makecell{ResNet-34}}
& CE Loss & 0.751$\pm0.007$ & 0.917 & 0.407 & 0.564 & 0.757 & 0.886 & 0.816 & 0.690 & 0.690 & 0.690 \\
& Poisson~\cite{beckham2017unimodal} & 0.701$\pm0.005$ & 0.606 & 0.740 & 0.667 & 0.811 & 0.682 & 0.741 & 0.600 & 0.714 & 0.652 \\
& NSB~\cite{liu2018ordinal} & 0.733$\pm0.006$ & 0.933 & 0.519 & 0.667 & 0.784 & 0.784 & 0.784 & 0.592 & 0.761 & 0.667 \\
& UDM~\cite{wu2019learning} & 0.707$\pm0.003$ & 0.642 & 0.333 & 0.439 & 0.753 & 0.795 & 0.773 & 0.640 & 0.762 & 0.696 \\
& CORF~\cite{zhu2021convolutional} & 0.764$\pm0.006$ & 0.923 & 0.444 & 0.600 & 0.726 & \underline{0.931} & 0.815 & \underline{0.838} & 0.619 & 0.712 \\
& MORF$_c$~\cite{lei2020meta} & 0.771$\pm0.003$ & \underline{1.000} & 0.518 & 0.683 & 0.839 & 0.772 & 0.804 & 0.629 & \textbf{0.928} & 0.750 \\
& \textbf{MORF (ours)} & \textbf{0.822$\pm0.004$} & 0.724 & \underline{0.778} & \textbf{0.750} & \textbf{0.880} & 0.830 & \textbf{0.854} & 0.778 & 0.833 & \textbf{0.805} \\
\midrule
\multirow{7}*{\makecell{VGG-16}}
& CE Loss & 0.688$\pm0.006$ & 0.571 & 0.148 & 0.235 & 0.710 & 0.863 & 0.779 & 0.651 & 0.667 & 0.658 \\
& Poisson~\cite{beckham2017unimodal} & 0.713$\pm0.005$ & 0.618 & \underline{0.778} & 0.689 & 0.861 & 0.636 & 0.732 & 0.603 & 0.833 & 0.700 \\
& NSB~\cite{liu2018ordinal} & 0.764$\pm0.005$ & \underline{1.000} & 0.370 & 0.540 & 0.781 & 0.852 & 0.815 & 0.686 & 0.833 & 0.752 \\
& UDM~\cite{wu2019learning} & 0.777$\pm0.004$ & 0.928 & 0.481 & 0.634 & 0.782 & 0.897 & 0.835 & 0.714 & 0.714 & 0.714 \\
& CORF~\cite{zhu2021convolutional} & 0.752$\pm0.003$ & 0.916 & 0.407 & 0.564 & 0.707 & \underline{0.989} & 0.824 & \underline{0.909} & 0.476 & 0.624 \\
& MORF$_c$~\cite{lei2020meta} & 0.796$\pm0.003$ & \underline{1.000} & 0.593 & 0.744 & 0.777 & 0.909 & 0.837 & 0.763 & 0.690 & 0.725 \\
& \textbf{MORF (ours)} & \textbf{0.854$\pm0.003$} & 0.807 & \underline{0.778} & \textbf{0.792} & \textbf{0.905} & 0.864 & \textbf{0.884} & 0.787 & \textbf{0.881} & \textbf{0.831} \\
\bottomrule
\end{tabular*}
\label{tab:busi}
\end{table*}

\begin{figure*}
\centering
 \includegraphics[width=0.8\linewidth]{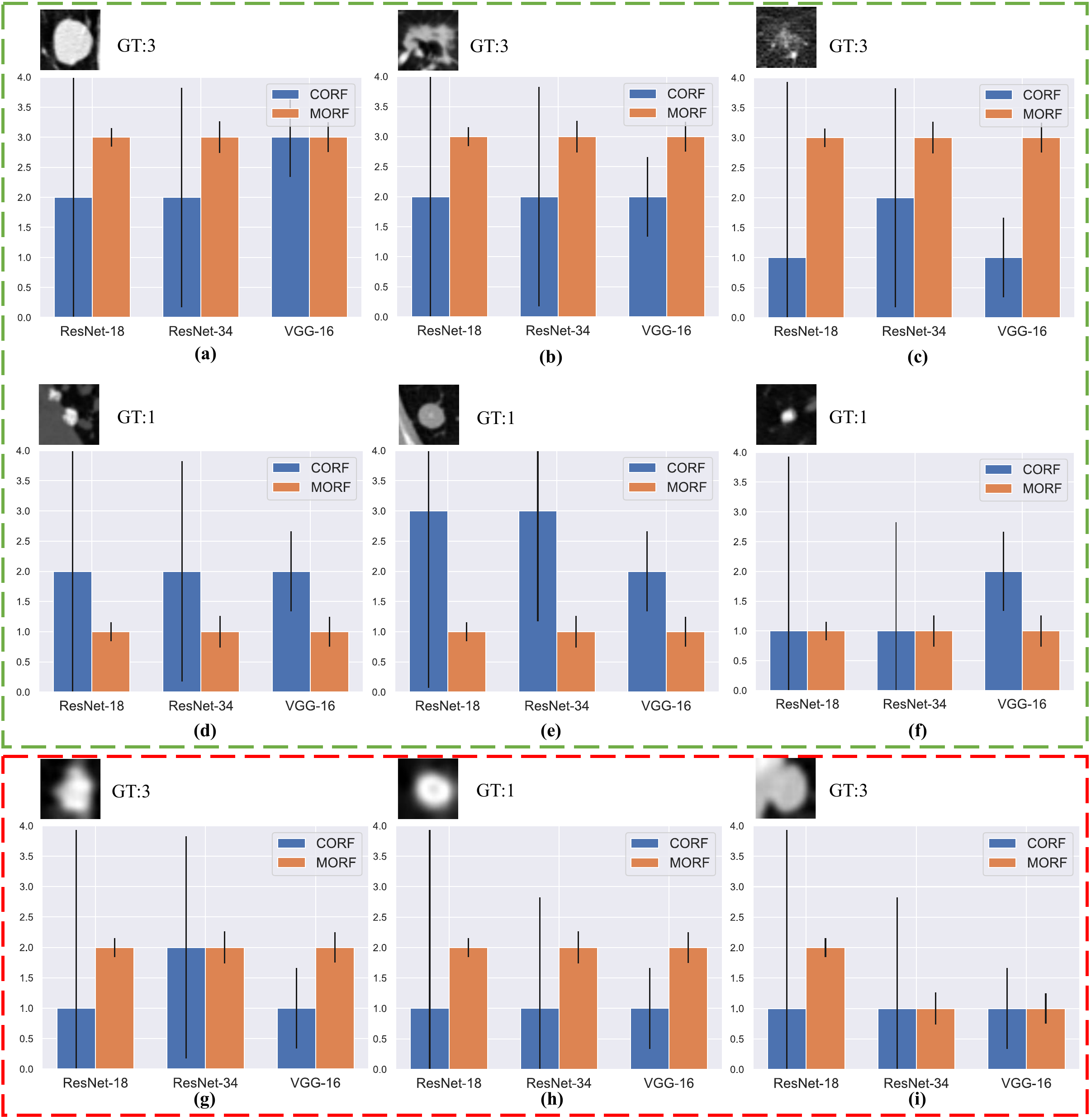}
\caption{Some prediction results of MORF and CORF under Train(3)-Test(2) setting. The $y$ axis represents the prediction results: 1, 2 and 3 represent benign, unsure and malignant. The GT denotes ground truth. The black vertical line on each bar represents the variance over the predictions of all trees. Green box contains some representative nodules, and red box include some failure cases.}
\label{fig:var}
\end{figure*}

\subsection{Training without Unsure Data for Lung Nodule Classification}
To verify the effectiveness of MORF on binary classification, we compare the results of all methods training without unsure data. The CE loss under Train(2)-Test(2) in Table~\ref{tab:train32_test2} (right) is the conventional binary classifier whose output dimension is $2$, and this is different from that of Train(3)-Test(2) whose output dimension is $3$. It is clear that the MORF also achieves the best overall accuracy, precision of benign, and recall of malignant using different backbones.

Through the comparison in Table~\ref{tab:train32_test2}, we can see that the unsure data largely affect the generalizability of the compared methods. There are no severe fluctuations in the performance of MORF under the two settings, indicating that MORF is able to distinguish the samples with ordinal labels regardless of whether the ordinal margin is large (without unsure) or small (with unsure). 

Here, we would like to clarify why the performance in Table~\ref{tab:train3_test3} is much lower than that seen in Table~\ref{tab:train32_test2}. This is due to the imperfect performance of recognizing the unsure samples, so that it becomes unavoidable to encounter the classification errors of all classes.

\subsection{Classification Results on BUSI Dataset}
In Table~\ref{tab:busi}, we illustrate the results of all the methods on the BUSI dataset. For the benign and malignant classes, we also focused more on the precision of benign and recall of malignant. Different from the LIDC-IDRI dataset, we can see from Fig.~\ref{fig:examples} that the first order in the BUSI dataset is the normal class, which does not contain nodules. Therefore, the benign class occupies different positions in the orders of the two datasets. Interestingly, our MORF also retains the best precisions of benign and malignant recalls as shown in Table~\ref{tab:busi}. This demonstrates the discriminative ability of MORF in recognizing nodules of different orders without the influence of the normal class that does not include nodules.

Clinically, false positives of the normal class indicate that benign or malignant nodules are falsely classified as normal, which will result in an increase of missing diagnosis; in contrast, false negatives of the normal class will cause an increase in misdiagnosis. Since the precision and recall correlate with false positives and the false negative, here we suggest that the precision and recall of the normal class have equal importance weights. The results of compared methods in Table~\ref{tab:busi} show that they are prone to preferring precision or recall of the normal class. For example, the UDM obtains a precision of $1.000$ when using ResNet-18 while the corresponding recall is $0.259$; the NSB achieves similar results with a large margin between precision and recall. However, the MORF has relatively balanced precisions and recalls, and it also maintains the best F1 scores when applying all of the backbones.

\subsection{Comparisons between MORF and MORF$_c$}
From Tables~\ref{tab:train3_test3},~\ref{tab:train32_test2}, and~\ref{tab:busi}, we can see that MORF consistently outperforms MORF$_c$ in terms of overall accuracy, which is benefited from the improved GFS using random selection without replacement. For both of the two datasets, MORF is better at identifying benign and malignant classes compared with MORF$_c$, and this guarantees the improvements of overall performance. Recalling the essential difference between MORF and MORF$_c$ that GFS in MORF uses random selection without replacement, as a result, MORF makes more efficient use of FC output vector while MORF$_c$ could be affected by selection of unused elements in FC output vector. Therefore, the results suggest that GFS without replacement will be more conducive to distinctive feature learning that identifies benign and malignant.

\subsection{Tree-wise Variance Reduction}

As analyzed in Section~\ref{subsec:optimization} that all trees in our MORF are influenced by the gradient of the meta loss through the TWW-Net weighting scheme. Here, we provide experiments to verify the consistency of behaviors of different trees. This experiment was conducted on LIDC-IDRI dataset. We estimated the variance over predictions of all trees: $var := \frac{1}{T} \sum_{t=1}^{T} |p_{t} - \hat{p}|^{2}$, where $p_{t}$ and $\hat{p}$ denote the prediction of the $t$-th tree and the final prediction, respectively. In Fig.~\ref{fig:var}, we compare the predictions and the corresponding $var$s on some special nodules such as cavities (\textbf{b}), ground-glass (\textbf{c}), calcifications (\textbf{d, f}) and benign nodule with larger sizes (\textbf{e}). For the prediction results, the MORF is more accurate than CORF on various kinds of nodules using different backbones. Especially for the ground-glass nodules and the large benign nodules, the CORF makes incorrect  predictions, which will cause severe diagnosis loss. For the large malignant nodule (\textbf{a}) and the calcification case (\textbf{d}), the CORF tends to regard them as unsure while delaying the treatment of the patient with the malignant nodule. This defect may be attributed to the lower feature randomness in CORF. Moreover, the $var$s of MORF are much lower than those of CORF in all cases. This indicates the behavioral consistency of all of the trees in MORF, which benefits from the guidance of meta GFS features and the meta training procedure.

In Fig.~\ref{fig:var}, we also provide some failure cases obtained by MORF and CORF, \ie, \textbf{(g)}, \textbf{(h)}, \textbf{(i)}. \textbf{(g)} is malignant and its real malignant score is $3.75$ which is referred to Fig.~\ref{fig:distribution}. However, both MORF and CORF make incorrect predictions, and this is due to blur edges or shapes of the nodule. \textbf{(h)} is a benign nodule in our study with malignant score $2.5$ which is an upper bound of score range of benign class. We can see that MORF is prone to classifying it as unsure, while CORF obtains predictions of benign. \textbf{(i)} is a malignant nodule with malignant score $3.5$, \ie, the lower bound of malignant, and the predictions are incorrect obtained by the two methods. We conclude that MORF and CORF can be confused by the malignant or benign nodules whose malignant scores are close to unsure class, and MORF prefers predictions of unsure for them. This phenomenon also reflects that MORF is more suitable for real clinical circumstance that requires more nodules surrounding the unsure class for further diagnosis.

\subsection{Ablation Study}
Here, we evaluated the effectiveness of the GFS module and TWW-Net based on CORF. The experiments were conducted on the LIDC-IDRI dataset under the Train(2)-Test(2) setting using the ResNet-18 backbone. Then, we evaluated the effects of the number and depth of trees in MORF. Finally, we conducted a significance analysis between MORF and other methods through predictions on the LIDC-IDRI testing set (Train(3)-Test(3)).
\begin{table}[h]
\caption{Effectiveness evaluation of GFS and TWW-Net on the LIDC-IDRI dataset, under the Train(2)-Test(2) setting using ResNet-18.}
\centering
\begin{tabular*}{1.0\linewidth}{@{\extracolsep{\fill}}l*{5}{c}}
\toprule
\multirow{2}{*}{Method} & \multirow{2}{*}{Accuracy} & \multicolumn{2}{c}{Benign} & \multicolumn{2}{c}{Malignant} \\ \cline{3-6}
 &  & Precision & F1 & Recall & F1 \\
\midrule
CORF & 0.830 & 0.893 & 0.874 & 0.772 & 0.739 \\
\quad+GFS & 0.843 & 0.881 & 0.886 & 0.733 & 0.744 \\
\quad+TWW-Net & 0.821 & 0.897 & 0.897 & 0.653 & 0.737 \\
\textbf{MORF} & \textbf{0.886} & \textbf{0.915} & \textbf{0.917} & \textbf{0.812} & \textbf{0.816} \\
\bottomrule
\end{tabular*}
\label{tab:ablation}
\end{table}
\subsubsection{GFS module}  To verify the random feature perturbation enforced by the GFS, we added the GFS to CORF termed CORF+GFS. The forest of the CORF is fixed during training and inference. In contrast, the CORF+GFS enables the forest structure to be dynamic during training only, and the training process does not include the meta train stage. That is to say, the CORF is equipped with random feature perturbations. From Table~\ref{tab:ablation} we can see that the CORF+GFS achieves better performances than the vanilla CORF. This indicates that the training of the CORF benefits from the GFS in that the GFS generated forest endows the target model ($\mat{\theta}$) with the generalizability increased by the random feature perturbation. However, the drawback of all trees sharing the same weights is still not resolved. In addition, we observe that the precision of benign and the recall of malignant of CORF+GFS are worse than CORF, which can be explained as follows: GFS is specially designed to improve the meta-training of the proposed MORF while CORF does not have meta training, as a result, the GFS shall compromise the performance of CORF as expected.

\subsubsection{TWW-Net}  The CORF+TWW-Net in Table~\ref{tab:ablation} is tailored for evaluating the TWW-Net without the GFS, \ie, the structure of the forest is also fixed, and the training process includes the meta train stage. Table~\ref{tab:ablation} shows that CORF+TWW-Net performs worse than the CORF. This is due to that the training data and the meta data are the same and consequently, the two terms of the multiplication in $G_{ij}$ as shown in~\eqref{eq:new_update_phi} are the same. Therefore, the $G_{ij}$ is almost at the orientation of the largest gradient, and this phenomenon happens equally to all training samples. So we argue that in~\eqref{eq:new_update_phi}, CORF+TWW-Net could accelerate the update of $\vct{\phi}$, and hence, may trigger the overfitting of TWW-Net. In other words, the update of $\vct{\theta}$ is not guided by the meta data. Consequently, the meta weighting scheme of TWW-Net should be driven by the model generalizability gain from GFS. For MORF (\ie~CORF + GFS + TWW-Net), the update of parameters $\vct{\theta}$ can be guided by GFS generated features, therefore, $G_{ij}$ involves gradients of GFS features, then $G_{ij}$ will slow down the update of $\vct\phi$ according to~\eqref{eq:new_update_phi}. Hence, the combination of GFS and meta training with TWW-Net achieves trade-off between updating the parameters $\vct{\theta}$ and $\vct{\phi}$.

\subsubsection{Number and depth of trees}

\begin{figure}[ht]
\centering
 \includegraphics[width=1.0\linewidth]{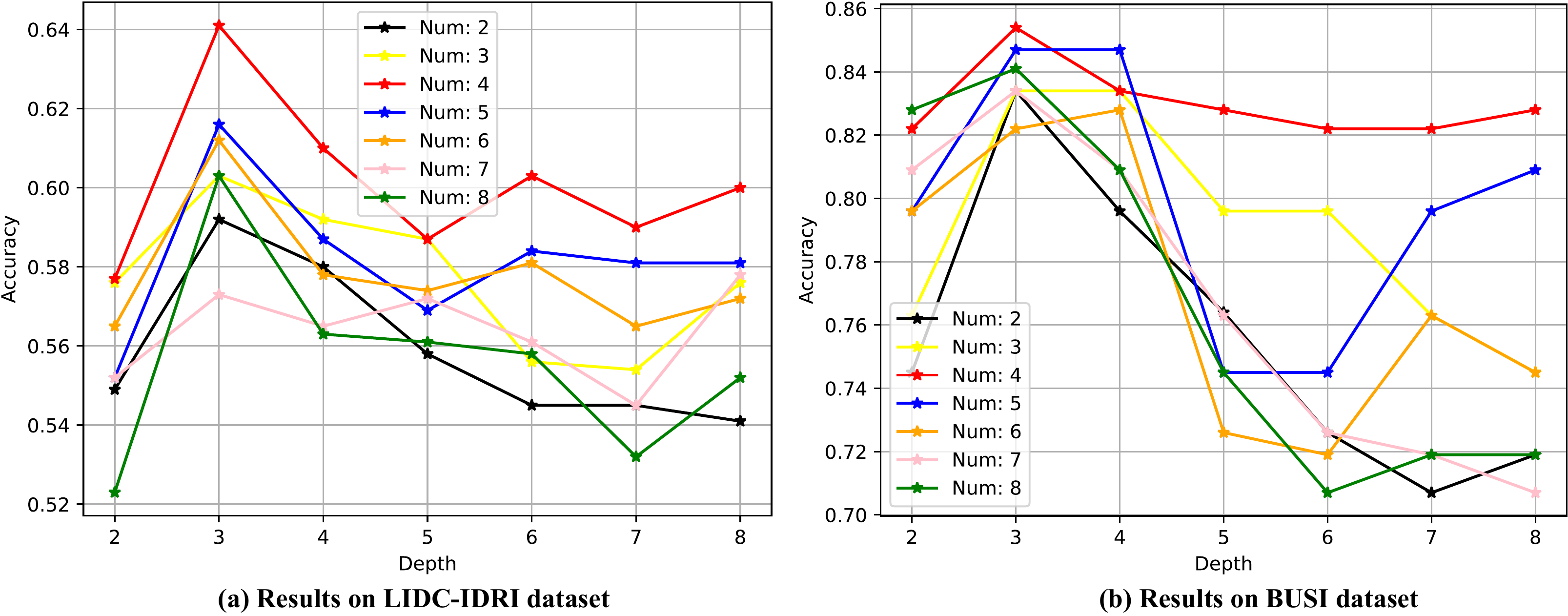}
\caption{Classification accuracies with varying values of the depth of trees and number of trees on \textbf{(a)} LIDC-IDRI and \textbf{(b)} BUSI datasets.}
\label{fig:number_depth}
\end{figure}

Here, we further discuss the effects of the number and depth of trees. We fix one of them and evaluate the settings with various values of another. The backbone network is VGG-16. Fig.~\ref{fig:number_depth} shows the performances influenced by these two factors on the LIDC-IDRI under the setting of Train(3)-Test(3) and BUSI datasets.

In Fig.~\ref{fig:number_depth}, we can see that the setting with the number of trees being 4 and the depth of trees being 3 achieves the best performances for both of datasets. Lower or higher values of number and depth will decrease the performances. If the number of trees is fixed, there will be more nodes in a forest which requires FC output vector to have higher dimension, \ie, more elements. Then the parameters of the framework begin to increase, and this will affect the performance. If the depth of trees is fixed, we observe that when the number is small, \ie, $2$ or $3$, MORF achieves lower performances on two datasets. This indicates that fewer ensembled trees can affect the capability of the MORF, then hinder the performance improvement. In contrast, too many trees ($>4$) also results in a decrease of performance which is attributed to more MLPs for weights learning, \ie, there are more parameters to learn.

\subsection{Significance Analysis}
In order to show the significant differences between MORF and compared methods, we compare the differences between MORF and other methods through conducting the Wilcoxon signed-rank test~\cite{wilcoxon} with respect to predicted probabilities on test set. In Table~\ref{tab:significance}, we can see that all the $p$-values obtained by MORF and baseline methods are less than the significance level $0.05$. So we reject the null hypothesis that MORF possesses the same prediction distribution as baseline methods. Finally, we conclude that MORF is \emph{significantly} better than the baseline methods.

\begin{table}
\scriptsize
\caption{$p$-values between MORF and other methods using different backbones.}
\centering
\begin{tabular*}{\linewidth}{lccccc}
\toprule
Backbone & CE Loss & Poisson~\cite{beckham2017unimodal} & NSB~\cite{liu2018ordinal} & UDM~\cite{wu2019learning} & CORF~\cite{zhu2021convolutional} \\
\midrule
ResNet-18 & $<$0.01 & $<$0.01 & $<$0.01 & $<$0.01 & $<$0.01 \\
ResNet-34 & $<$0.01 & $<$0.01 & $<$0.01 & $<$0.01 & $<$0.01 \\
VGG-16 & $<$0.01 & $<$0.01 & $<$0.01 & $<$0.01 & $<$0.01 \\
\bottomrule
\end{tabular*}
\label{tab:significance}
\end{table}

\section{Conclusion}
\label{sec:con}
In this paper, we propose a meta ordinal regression forest, termed MORF, for improving the performances of the ordinal regression in medical imaging, such as lung nodule classification and breast cancer classification. The MORF contains a grouped feature selection module that is used to generate a dynamic forest with feature random perturbation. Another critical component of the MORF is the TWW-Net which assigns each tree with a learned weight, and this enforces the predictions of different trees to have smaller variance while maintaining stable performances. The parameters of the model are learned through the meta-learning scheme which can solve the problem of integrating two parts of parameters into one training loop, and it brings the gradients of target data and meta data to be closer. Through the experiments, we have verified that the MORF can help reduce the false positives and the false negatives of the relevant classes, which is significant to the clinical diagnosis. Moreover, we have also verified that the accurate recognition of the intermediate order can improve the classification of the other classes on both sides. 

In the future, we will consider to explore attention mechanism for achieving better generalizability of deep random forest such as in~\cite{xuan2018refined}. Also, we will simplify the MORF framework such as using the design of light weight network or efficient loss functions~\cite{liang2021cemodule,qin2020pairwise}.


\ifCLASSOPTIONcaptionsoff
  \newpage
\fi

\end{document}